\documentclass[sigconf]{acmart}
\AtBeginDocument{%
  \providecommand\BibTeX{{%
    \normalfont B\kern-0.5em{\scshape i\kern-0.25em b}\kern-0.8em\TeX}}}

\usepackage{graphicx}
\usepackage{amsmath}
\usepackage{amssymb}
\usepackage{amsthm}
\usepackage{booktabs}
\usepackage{algorithm}
\usepackage{algorithmic}
\usepackage{dsfont}
\usepackage{multirow}
\usepackage{bm}
\usepackage{pifont}
\usepackage{bbding}
\newcommand{\tabincell}[2]{\begin{tabular}{@{}#1@{}}#2\end{tabular}}

\newcommand\blfootnote[1]{% 
\begingroup 
\renewcommand\thefootnote{}\footnote{#1}% 
\addtocounter{footnote}{-1}% 
\endgroup 
}

\setcopyright{acmcopyright}

\copyrightyear{2022}
\acmYear{2022}
\setcopyright{acmcopyright}\acmConference[MM '22]{Proceedings of the 30th ACM
International Conference on Multimedia}{October 10--14, 2022}{Lisboa, Portugal}
\acmBooktitle{Proceedings of the 30th ACM International Conference on Multimedia
(MM '22), October 10--14, 2022, Lisboa, Portugal}
\acmPrice{15.00}
\acmDOI{10.1145/3503161.3547969}
\acmISBN{978-1-4503-9203-7/22/10}

\begin{document}
% \fancyhead{}
%%
%% The "title" command has an optional parameter,
%% allowing the author to define a "short title" to be used in page headers.
\title{Reducing the Vision and Language Bias for Temporal Sentence Grounding}

\author{Daizong Liu$^*$}
\email{dzliu@stu.pku.edu.cn}
\affiliation{%
  \institution{Wangxuan Institute of Computer \\
Technology, Peking University}
  \city{Beijing}
  \country{China}
  }
  
\author{Xiaoye Qu$^*$}
\email{quxiaoye@huawei.com}
\affiliation{%
  \institution{Huawei Cloud}
  \city{Hangzhou}
  \country{China}
  }

\author{Wei Hu$^\dagger$}
\email{forhuwei@pku.edu.cn}
\affiliation{%
  \institution{Wangxuan Institute of Computer \\
Technology, Peking University}
  \city{Beijing}
  \country{China}
  }

\renewcommand{\shortauthors}{Daizong Liu \& Xiaoye Qu \& Wei Hu}
%%
%% By default, the full list of authors will be used in the page
%% headers. Often, this list is too long, and will overlap
%% other information printed in the page headers. This command allows
%% the author to define a more concise list
%% of authors' names for this purpose.
% \renewcommand{\shortauthors}{Trovato and Tobin, et al.}

%%
%% The abstract is a short summary of the work to be presented in the
%% article.
\begin{abstract}
Temporal sentence grounding (TSG) is an important yet challenging task in multimedia information retrieval.
% \textcolor{red}{Due to lacking in sufficient pairwise video-query data for training},
Although previous TSG methods have achieved decent performance, they tend to capture the selection biases of frequently appeared video-query pairs in the dataset rather than present robust multimodal reasoning abilities, especially for the rarely appeared pairs. 
In this paper, we study the above issue of selection biases
% named as few-shot temporal sentence grounding, 
and accordingly propose a Debiasing-TSG (D-TSG) model to filter and remove the negative biases in both vision and language modalities for enhancing the model generalization ability.
Specifically, we propose to alleviate the issue from two perspectives:
1) Feature distillation. We built a multi-modal debiasing branch to firstly capture the vision and language biases, and then apply a bias identification module to explicitly recognize the true negative biases and remove them from the benign multi-modal representations.
2) Contrastive sample generation. We construct two types of negative samples to enforce the model to accurately learn the aligned multi-modal semantics and make complete semantic reasoning.
We apply the proposed model to both commonly and rarely appeared TSG cases, and demonstrate its effectiveness by achieving the state-of-the-art performance on three benchmark datasets (ActivityNet Caption, TACoS, and Charades-STA).
\end{abstract}

%%
%% The code below is generated by the tool at http://dl.acm.org/ccs.cfm.
%% Please copy and paste the code instead of the example below.
%%


%%
%% Keywords. The author(s) should pick words that accurately describe
%% the work being presented. Separate the keywords with commas.

%% A "teaser" image appears between the author and affiliation
%% information and the body of the document, and typically spans the
%% page.
% \begin{teaserfigure}
%   \includegraphics[width=\textwidth]{sampleteaser}
%   \caption{Seattle Mariners at Spring Training, 2010.}
%   \Description{Enjoying the baseball game from the third-base
%   seats. Ichiro Suzuki preparing to bat.}
%   \label{fig:teaser}
% \end{teaserfigure}

%%
%% This command processes the author and affiliation and title
%% information and builds the first part of the formatted document.
\maketitle

\blfootnote{
$^*$Equal Contribution.\\
$^\dagger$This work is supported by the National Key R\&D Program of China under contract No. 2021YFF0901502.
Corresponding author: Wei Hu (forhuwei@pku.edu.cn).}

\section{Introduction}
\begin{figure}[t!]
\centering
\includegraphics[width=0.47\textwidth]{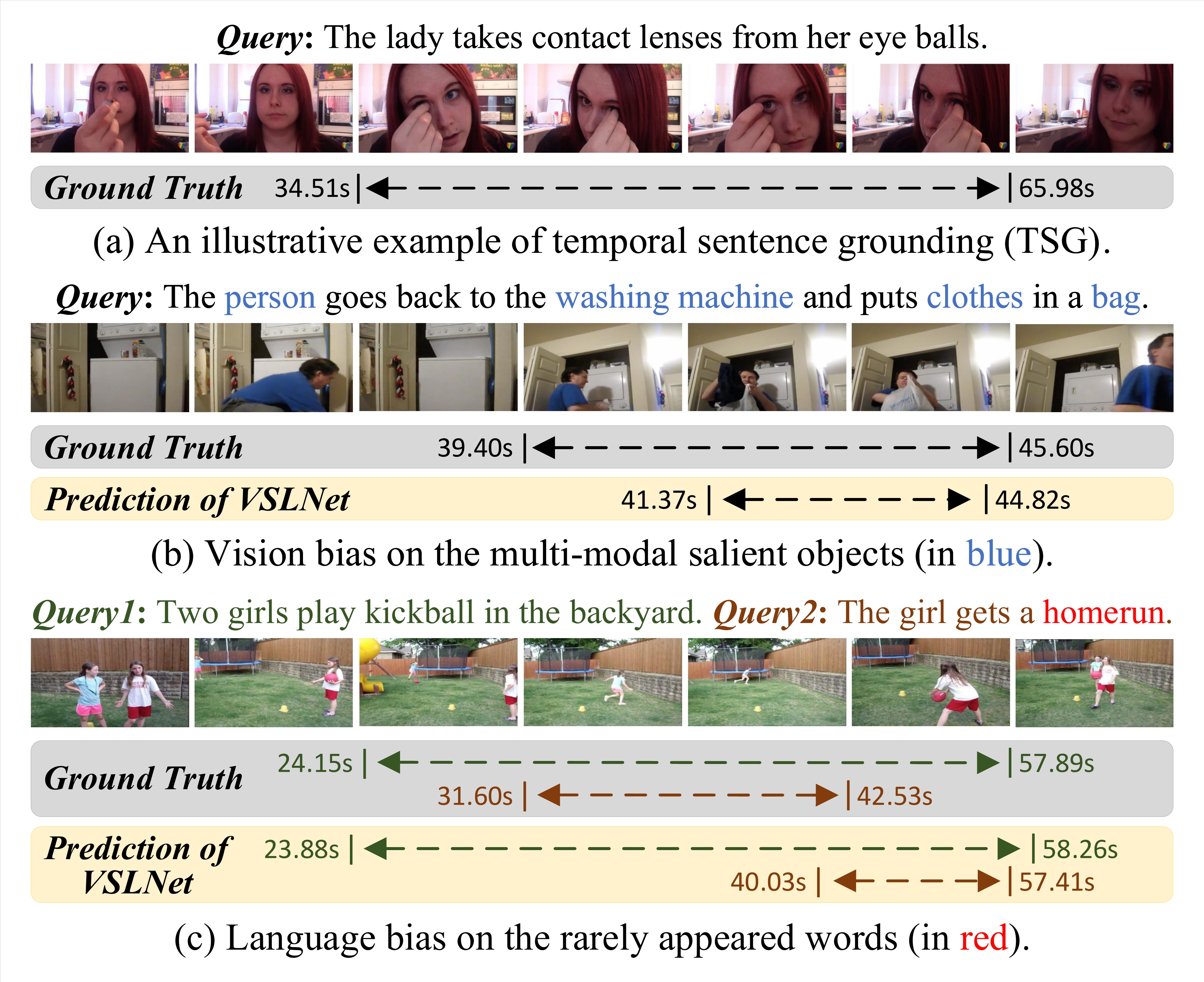}
\vspace{-8pt}
\caption{(a) The TSG task aims to localize a video segment semantically according to the query. (b) Examples of vision bias: existing methods predict the segment usually corresponding to the most query-related salient objects/attributes in the video. (c) Examples of language bias: existing methods usually learn to localize the commonly appeared words well, but fail to localize the rarely appeared ones.}
\label{fig:introduction}
\vspace{-18pt}
\end{figure}

Vision-and-language understanding is a fundamental task in multimedia field and has attracted increasing attention over the last years due to its various applications in video summarization \cite{song2015tvsum,chu2015video}, video captioning \cite{jiang2018recurrent,chen2020learning}, and temporal action localization \cite{shou2016temporal,zhao2017temporal}, etc.
Recently, temporal sentence grounding (TSG) \cite{gao2017tall,anne2017localizing} has been proposed as an important yet challenging task. This task requires automatically determining the start and end timestamps of a target segment in an untrimmed video that contains an activity semantically related to a given sentence description, as shown in Figure \ref{fig:introduction} (a). It needs to not only model the complex multi-modal interactions among vision and language features, but also capture complicated context information for their semantic alignment.

To localize the target segment, most previous works  \cite{anne2017localizing,gao2017tall,chen2018temporally,zhang2019cross, liu2020jointly,zhang2019learning,liu2020reasoning,liu2021adaptive,liu2021progressively,liu2022exploring,liu2022skimming} first pre-define abundant video segments as proposals, and then rank them by matching with the sentence query. The best segment proposal with the highest matching score is finally selected as the target segment. Instead of proposal-based paradigm, some proposal-free works \cite{rodriguez2020proposal,chenrethinking,yuan2019find,mun2020local,zhang2020span,liu2022unsupervised} propose to directly regress the start and end timestamps of the target segment for each frame. 
These methods are more efficient and achieve faster running speed than the counterparts using segment proposals.

Although the above two types of methods have achieved impressive results, some recent studies \cite{nan2021interventional,liu2022memory} indicate that previous TSG models suffer from the selection bias existing in the dataset. That is, existing models tend to localize the queries that occur frequently in the dataset but cannot generalize well to the rarely appeared queries. To reduce this tendency, during both training and inference phases,
\cite{nan2021interventional} introduces an additional interventional causal model and \cite{liu2022memory} utilizes large number of memory parameters to mitigate the spurious correlations guided by the rare queries.
% \cite{nan2021interventional,liu2022memory} introduced additional large number of parameters in both training and inference phase to improve model generalization. 
% \textcolor{red}{deatiled description of nan2021interventional and liu2022memory}. 
However, both works ignore the implicit vision bias and their performance are still under satisfactory.
Different from them, in this paper, we attempt to effectively reduce both the vision and language bias \textit{without} introducing any additional parameters in inference phase.

Formally, we denote the vision bias as that models converge into the partial semantic matching, which only activates the most salient objects/attributes in the video instead of matching the full semantics of the sentence. As shown in Figure~\ref{fig:introduction} (b), previous method localizes the segment with all objects appearing but fails to capture the segment describing action ``goes back". 
Therefore, alleviating the negative effect of vision bias is crucial for learning a complete correspondence between visual contents and sentence semantics. 
Moreover, Figure \ref{fig:introduction} (c) depicts the language bias where the TSG models fail to localize the query with rare words (\textit{e.g.}, homerun) well. Thus, it is also necessary to relieve such language bias for improving model generalization. 
% Since some ``biases" captured from the dataset may represent the natural rule in real world (\textit{i.e.}, commonsense knowledge), they are not harmful to the models and models may benefit from them. For example, ``whale" is a kind of ``animal". Thus, it is also necessary to detect and filter the true negative biases in vision and language.

To this end, we propose a novel model named Debiasing-TSG (D-TSG) for TSG task, to overcome the negative vision-language biases from two perspectives including feature distillation and contrastive sample generation.
Specifically, 1) from the feature perspective, we build a multi-modal debiasing branch to capture the vision and language biases, respectively.
% Since some ``biases" captured from the dataset may represent the natural rule in real world (\textit{i.e.}, commonsense knowledge), they are not harmful to the models and models may benefit from them. For example, ``whale" is a kind of ``animal". 
However, we argue that not all these biases captured from the dataset are harmful to the TSG models as some important nouns, verbs, and video features also contribute to accurate localization. 
Therefore, a bias identification module is further devised to filter and remove the true negative biases in both two modalities. 
Besides, to maintain the inference efficiency, we do not expect to introduce additional parameters in the inference phase. To achieve it, we adopt a contrastive learning process to approach the multi-modal features to the debiased features as close as possible. In this way, in the inference phase, we can simply remove our developed debiasing branch, and directly adopt the benign multi-modal features as it have a similar representation as the debiased features.
2) From the sample perspective, the TSG model should make full use of the information of queries and videos for improving the generalization of the model. Therefore, for each video-query pair in the dataset, we construct two types of negative samples to assist the model learning the complete semantics in both vision and language modalities.

Our contributions can be summarised as follows:
\vspace{-10pt}
\begin{itemize}
    \item In this paper, we study both the vision and language bias existing in the previous TSG models and propose a novel D-TSG model to reduce them. To the best of our knowledge, it is the first time that a framework is proposed to detect and remove the bias in both vision and language modalities.
    \item We develop a D-TSG model from two perspectives of feature distillation and contrastive sample generation. Our model is both effective and efficient, and we do not introduce any additional parameters to the backbone TSG model in the inference phase.
    \item The proposed D-TSG achieves state-of-the-art performance on three benchmarks (ActivityNet Caption, TACoS, and Charades-STA), improving the performance by a large margin not only on the entire dataset but also on the rarely appeared pairwise samples.
\end{itemize}

% \vspace{10pt}
\section{Related Work}
\noindent \textbf{Temporal sentence grounding.}
The task of temporal sentence grounding (TSG) is introduced by \cite{gao2017tall} and \cite{anne2017localizing}, which aims to identify the precise start and end timestamps of one specific video segment semantically corresponding to the given sentence query. 
The early works \cite{anne2017localizing,ge2019mac,liu2018attentive,zhang2019man,chen2018temporally,zhang2019cross,liu2018cross,yuan2019semantic,xu2019multilevel,liu2020jointly} employ a proposal-based architecture that first generates segment proposals, and then ranks them according to the similarity between proposals and the query to select the best matching one. Some of them \cite{gao2017tall,anne2017localizing} propose
to apply the sliding windows to generate proposals and subsequently integrate the query with segment representations via a matrix operation. 
Instead of using the sliding windows, latest works \cite{wang2019temporally,zhang2019man,yuan2019semantic,zhang2019cross} directly integrate sentence information with each fine-grained video clip unit, and predict the scores of candidate segments by gradually merging the fusion feature sequence over time. Although those methods achieve promising performances, they are severely limited by the quality of proposals and the large amount of proposals leads to poor inference efficiency.
To overcome above drawback, recent works \cite{chenrethinking,yuan2019find,mun2020local,zhang2020span,nan2021interventional} follow a proposal-free paradigm that directly regresses the temporal locations of the target segment.
% They do not rely on the segment proposals and directly select the starting and ending frames by leveraging cross-modal interactions between video and query.
Specifically, they either regress the start/end timestamps based on the entire video representation \cite{yuan2019find,mun2020local}, or predict at each frame to determine whether this frame is a start or end boundary \cite{zhang2020span,nan2021interventional}.
Therefore, these methods perform segment localization more efficiently than the proposal-based ones.

\begin{figure*}[t!]
\centering
\includegraphics[width=1.0\textwidth]{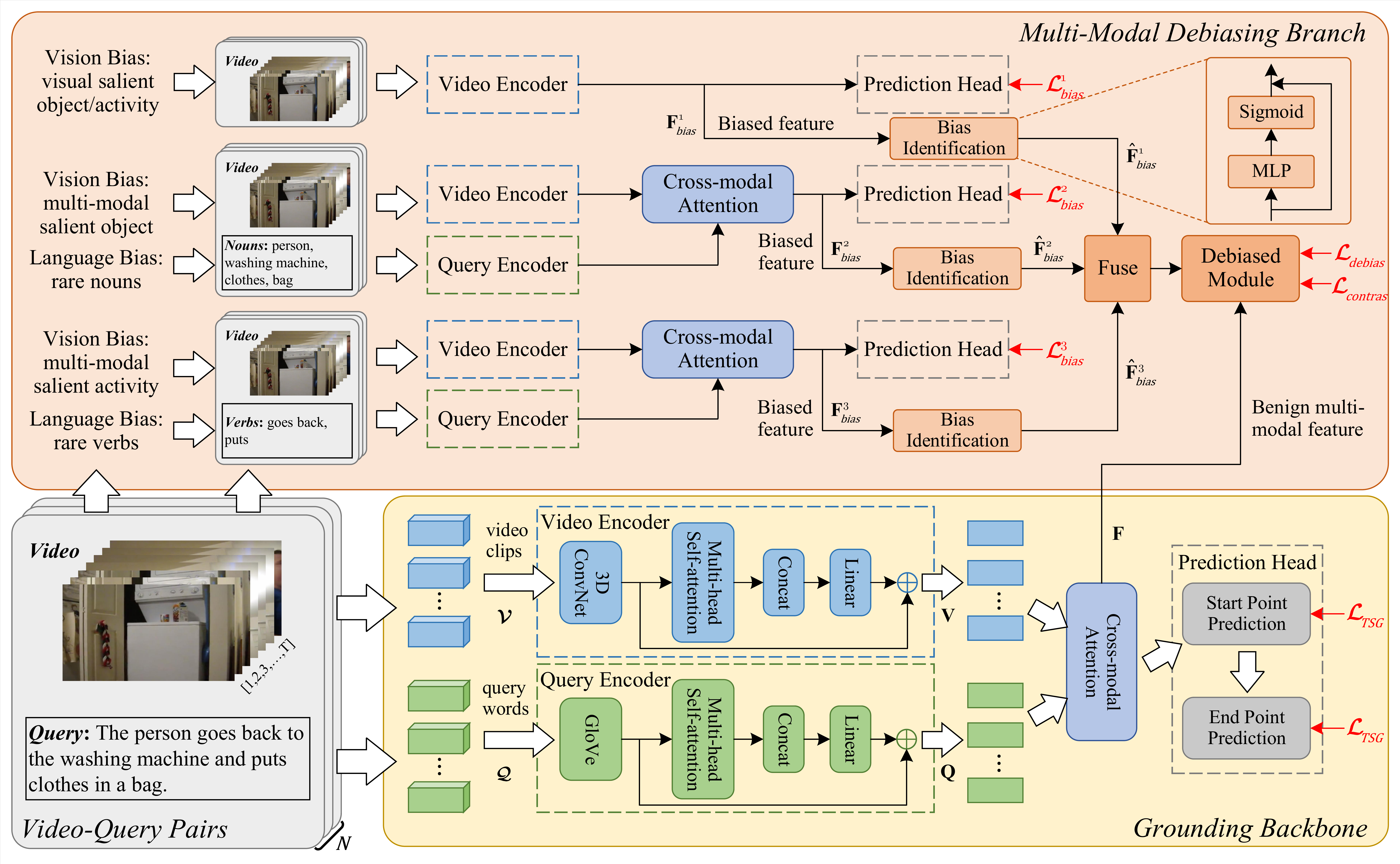}
% \vspace{-14pt}
% \vspace{-10pt}
\caption{Overview of our proposed D-TSG model. We only illustrate the debiasing from the feature perspective. Given paired video-query data, the grounding backbone is utilized to directly predict the target segment during inference. During training stage, we additionally develop a multi-modal debiasing branch to detect and reduce the negative vision and language biases.}
\label{fig:pipeline}
\vspace{-4pt}
\end{figure*}

However, the above two types of methods are limited by the bottleneck of the selection bias \cite{nan2021interventional,liu2022memory} in the dataset.
That is, existing TSG models tend to forget the rare cases while learning on a dataset distributed off-balance and diverse, especially in practical scenarios where the data distribution could be extremely imbalanced.
Although \cite{nan2021interventional} develop an additional causal model and \cite{liu2022memory} employ additional memory bank to make the complete semantic learning on the rare query, these works only focus on the language bias while ignoring the crucial vision bias during the model learning. Besides, they also introduce additional large number of parameters in the inference phase, degenerating the network inference efficiency.
Different from them, in this paper, we consider alleviating the biases in both vision and language modalities without introducing any parameters during inference stage.
% \noindent \textbf{Few-shot learning.}
% Conventional few-shot learning \cite{ren2018meta,wang2018zero,xian2016latent} usually focuses on single-label classification. Other researchers \cite{lee2018multi,fu2015transductive} further study the few-shot problem in the context of multilabel classification. Hendricks \textit{et al.} \cite{hendricks2016deep,venugopalan2017captioning} propose the task of few-shot image captioning, which can be regarded as sentence classification. In addition to the few-shot classification, there are also many work focusing on few-shot vision-language matching. Socher \textit{et al.} \cite{socher2013zero} and Frome \textit{et al.} \cite{frome2013devise} use visual-semantic matching frameworks to recognize unseen objects in images. 
% % Different from them, we aim to deal with the few-shot contents for TSG task. 
% Rather than single or multiple words, here we aim to deal with the few-shot contents for TSG task, which include not only multiple few-shot words but also other few-shot phrases, as well the video contents.
% Besides, our training stage is the same as previous TSG methods that utilize all video-query pairs in train set for training, the only difference is that we split the test data into rare and common sets for evaluation.

\noindent \textbf{Contrastive learning.}
Contrastive learning \cite{van2018representation} aims to learn the high-level representation by maximising the mutual information between the input samples and positive samples, which has been applied in many fields \cite{liang2020learning,qiu2021source,zhang2021unleashing}. Specifically, some contrastive learning methods \cite{chen2020simple,he2020momentum} adopt contrastive learning to train the models by decreasing the distance between the feature representations of different augmented views of the same images, while increasing the distance between different images. Liang \textit{et al.} \cite{liang2020learning} adopt contrastive learning to approach the feature representations of the original samples to the factual samples, while keeping away from that between the original and counterfactual samples. In this paper, we adopt contrastive learning in both two parts: 1) For feature distillation without introducing additional parameters, we use contrastive learning to push the multi-modal feature to approach the debiased features, and meanwhile to far away from the negative biased features. 2) For contrastive samples, we use contrastive learning to increase the sensitivity and generalization of the model.

\section{The Debiasing-TSG Model}
\subsection{Overview}
Given an untrimmed video $\mathcal{V}$ and a sentence query $\mathcal{Q}$, the task aims
to determine the start and end timestamps $(s,e)$ of a specific video segment referring to the sentence query.
Formally, we represent the video as $\mathcal{V}=\{v_t\}_{t=1}^T$ clip-by-clip, where $v_t$ is the $t$-th clip in the video and $T$ is the total clip number. 
We also denote the given sentence query as $\mathcal{Q}=\{q_j\}_{j=1}^M$ word-by-word, where $q_j$ is the $j$-th word and $M$ is the total word number. With the whole dataset $\{\mathcal{V},\mathcal{Q},(s,e)\}^N$ of $N$ triplets, we aim to alleviate the negative effect of both vision and language  biases learnt from the dataset for enhancing the model generalization ability.

We present our method D-TSG in Figure~\ref{fig:pipeline}. As shown in this figure, we first design a grounding backbone model, which consists of a video encoder, a query encoder, a cross-modal interaction module, and a final prediction head. To reduce the negative biases learned in this backbone model, we construct a multi-modal debiasing branch to capture the vision and language biases. Especially, we devise a bias identification module and a debiased module in this branch to explicitly recognize and remove the true negative biases in two modalities. During the network training, we obtain the debiased features by removing the negative biased features from the benign multi-modal features of the grounding backbone, and adopt a contrastive learning strategy to approach the benign multi-modal features to the debiased features while keeping away from the negative biased features. In this way, in the inference, instead of introducing additional parameters, we can discard the developed multi-modal debiasing branch and directly utilize the grounding backbone to obtain more accurate multi-modal features for better grounding results. Besides, we also construct two types of contrastive samples to assist the model training.

\subsection{The Grounding Backbone}
\noindent \textbf{Video encoder.}
For video encoding, following previous works \cite{zhang2019cross,liu2020jointly}, we first extract the clip-level features by a pre-trained C3D network \cite{tran2015learning}, and then employ a multi-head self-attention \cite{vaswani2017attention} module to capture the long-range dependencies among video clips. We denote the extracted video features as $\bm{V}=\{\bm{v}_t\}^{T}_{t=1} \in \mathbb{R}^{T \times D}$, where $D$ is the feature dimension.

\noindent \textbf{Query encoder.}
For query encoding, following previous works \cite{liu2020jointly,qu2020fine}, we first obtain the word-level embeddings by the Glove model \cite{pennington2014glove}. We also apply another multi-head self-attention module to integrate the sequential information among the words. The final feature of the input query is denoted as $\bm{Q} = \{\bm{q}_j\}_{j=1}^M \in \mathbb{R}^{M \times D}$.

\noindent \textbf{Cross-modal interaction.}
After obtaining the encoded features $\bm{V},\bm{Q}$, we utilize a simple but effective co-attention mechanism \cite{lu-etal-2019-debug,chenrethinking} to capture the fine-grained cross-modal interactions between video and query features for semantic alignment. Specifically, we first calculate the similarity scores between $\bm{V}$ and $\bm{Q}$ as:
\begin{equation}
    \bm{S} = \bm{V}(\bm{Q}\bm{W}_S)^{\text{T}} \in \mathbb{R}^{T\times M},
\end{equation}
where $\bm{W}_S \in \mathbb{R}^{D\times D}$ projects the query features into the same latent space as the video. Then, we compute two attention weights as:
\begin{equation}
    \bm{A} = \bm{S}_r (\bm{Q}\bm{W}_S) \in \mathbb{R}^{T\times D}, \ \bm{B} = \bm{S}_r \bm{S}_c^{\text{T}} \bm{V} \in \mathbb{R}^{T\times D},
\end{equation}
where $\bm{S}_r$ and $\bm{S}_c$ are the row- and column-wise softmax results of $\bm{S}$, respectively. We compose the final query-guided video representation (\textit{i.e.}, multi-modal features) as follows:
\begin{equation}
    \bm{F} = \text{FFN}([\bm{V};\bm{A};\bm{V}\odot \bm{A};\bm{V}\odot \bm{B}]) \in \mathbb{R}^{T\times D},
\end{equation}
where $\bm{F}=\{\bm{f}_t\}^{T}_{t=1}$, $\text{FFN}(\cdot)$ denotes the feed-forward layer, $[;]$ is the concatenate operation, and $\odot$ is the element-wise multiplication.

\noindent \textbf{Prediction head.}
To predict the grounding results with the multi-modal features $\bm{F}=\{\bm{f}_t\}^{T}_{t=1}$, we employ the efficient proposal-free prediction head to regress the start and end timestamps of the segment. Specifically, we utilize two separate LSTM layers to successively predict the start and end scores on each video clip as:
\begin{equation}
\label{score1}
    \bm{h}^s_t = \text{LSTM}_{\text{start}}(\bm{f}_t,\bm{h}^s_{t-1}), \ C^s_t = [\bm{f}_t;\bm{h}^s_t] \bm{W}_s + \bm{b}_s,
\end{equation}
\begin{equation}
\label{score2}
    \bm{h}^e_t = \text{LSTM}_{\text{end}}(\bm{f}_t,\bm{h}^e_{t-1}), \ C^e_t = [\bm{f}_t;\bm{h}^e_t] \bm{W}_e + \bm{b}_e,
\end{equation}
where $\bm{h}$ is the hidden state of LSTM layer, $C^s_t,C^e_t$ denote the scores of start and end boundaries at $t$-th clip. We utilize the cross-entropy loss function $\mathcal{L}_{ce}$ to supervise this prediction head as:
\begin{equation}
    \mathcal{L}_{TSG} = \frac{1}{2T} \sum_{t=1}^T [\mathcal{L}_{ce}(C^s_t,\widehat{C}^s_t)+\mathcal{L}_{ce}(C^e_t,\widehat{C}^e_t)],
\end{equation}
where $\widehat{C}^s_t,\widehat{C}^e_t$ are the ground-truth labels. During the inference, we directly construct the top-n segments by considering the summed scores of the selected start and end boundary timestamps.

\subsection{Reducing the Bias by Feature Distillation}
In this part, we seek to reduce the negative effect of the bias from the feature perspective, namely, decreasing the negative biases from the multi-modal features $\bm{F}$. Specifically, we first need to capture the bias in both vision and language modalities, and then identify the true negative bias for the feature distillation.

\noindent \textbf{Capturing the vision-language bias.}
% To capture the vision and language bias learned from a TSG dataset, 
% one intuitive way is to train a biased model as a branch that takes the out-of-distribution feature as input. Specifically, 
We construct three biased models in the multi-modal debiasing branch to capture vision and language biases learned from a TSG dataset. As shown in Figure~\ref{fig:pipeline}, from top to bottom, we arrange these models from three aspects:
1) Most TSG methods tend to predict the segment corresponding to the most salient objects/activity in the video. For example, the videos in the TACoS dataset are all from cooking scenarios. At this time, we should reduce the bias from salient cooking activity to better identify target activity. To achieve this, we directly feed the single-modal video feature to predict the target segment for training and capture such biased feature $\bm{F}_{bias}^1 \in \mathbb{R}^{T\times D}$.
2) Besides, some TSG methods also tend to predict the segment only based on the query-related objects (\textit{i.e.}, nouns. words/phrases in the sentence) and thus fail to reason the complete query semantic. 
To capture such vision bias while exploring the language bias of the rare nouns, we first utilize the NLP tool \textit{spaCy} \cite{honnibal2017natural} to parse nouns from the query and subsequently concatenate them to construct a specific noun-based query. Then, we feed this specific query with the matched video to predict the segment for capturing the biased multi-modal feature $\bm{F}_{bias}^2 \in \mathbb{R}^{T\times D}$.
3) Similarly, some methods tend to predict the segment corresponding to the query-related actions (\textit{i.e.}, verbs. words/phrases in the sentence). Thus, we also utilize \textit{spaCy} \cite{honnibal2017natural} to parse verbs from the query, then feed them with the matched video to predict the segment for capturing the multi-modal feature $\bm{F}_{bias}^3 \in \mathbb{R}^{T\times D}$. 
% \textcolor{red}{Note that, the potential language bias of rare verbs can also be captured in this biased model.}
Specifically, we utilize the same video/query encoder, cross-modal interaction, and prediction head of the grounding backbone to train the above three biased models. The main difference to the grounding backbone is that we send the different inputs to the biased models for learning the biased features. We employ three cross-entropy losses $\mathcal{L}_{bias}^1,\mathcal{L}_{bias}^2,\mathcal{L}_{bias}^3$ to supervise the biased features $\bm{F}_{bias}^1,\bm{F}_{bias}^2,\bm{F}_{bias}^3$ learning process, respectively.

\noindent \textbf{Identifying the true negative bias.}
After obtaining the biased features $\bm{F}_{bias}^1,\bm{F}_{bias}^2,\bm{F}_{bias}^3$, we argue that not all these biases are harmful to the TSG models as some important nouns, verbs, and video features also contribute to accurate localization. 
% Some ``biases” captured and learned from the dataset contain commonsense knowledge (\textit{e.g.}, grouping the semantic of rare word ``whale" to ``animal") that may be helpful for the TSG models. 
Therefore, we devise a simple but effective bias identification module to identify the true negative biases that need to be removed in the benign multi-modal features $\bm{F}$.
Formally, our bias identification module is constructed by an MLP layer and a sigmoid activation function, which can be formulated as:
\begin{equation}
    \widehat{\bm{F}}_{bias}^i = \bm{F}_{bias}^i \cdot \text{sigmoid}(\text{MLP}(\bm{F}_{bias}^i)), \ i \in \{1,2,3\},
\end{equation}
where $\widehat{\bm{F}}_{bias}^i,i \in \{1,2,3\}$ are three detected negative bias.
To further fuse these multi-modal bias for obtaining the global negative biased feature $\widetilde{\bm{F}}_{bias}$, we adopt an attention mechanism as follows:
\begin{equation}
    \bm{m} = \text{softmax}(\text{MLP}([\widehat{\bm{F}}_{bias}^1;\widehat{\bm{F}}_{bias}^2;\widehat{\bm{F}}_{bias}^3])),
\end{equation}
\begin{equation}
    \widetilde{\bm{F}}_{bias} = \bm{m}^{\top}[\widehat{\bm{F}}_{bias}^1;\widehat{\bm{F}}_{bias}^2;\widehat{\bm{F}}_{bias}^3],
\end{equation}
where MLP is another MLP layer, $[;]$ is the concatenate operation.

\noindent \textbf{Obtaining the debiased feature.}
Intuitively, the debiased features can be obtained by directly removing the negative biased features $\widetilde{\bm{F}}_{bias}$ from the multimodal features $\bm{F}$, which can be represented as $\bm{F}_{debiased}=\bm{F}-\widetilde{\bm{F}}_{bias}$. 
With the feature $\bm{F}_{debiased}$ as input, we introduce a debiased module that contains an MLP layer and {a prediction head to supervise the debiasing process} with the loss function $\mathcal{L}_{debias}$.

% Note that, this debiased module is learned to produce the debiased features only in the training procedure since we aim to introduce no additional parameters in the inference phase.
To further introduce no additional parameters in the inference phase, we need to make the benign multi-modal features $\bm{F}$ generated from the grounding backbone and the debiased features $\bm{F}_{debiased}$ generated from the debiased branch as similar as possible in the training phases.
To this end, inspired by the contrastive learning strategy, we consider forcing the multi-modal features $\bm{F}=\{\bm{f}_t\}^{T}_{t=1}$ to near the debiased features $\bm{F}_{debiased}=\{\bm{f}_{debiased,t}\}^{T}_{t=1}$ and away from the negative biased features $\widetilde{\bm{F}}_{bias}=\{\widetilde{\bm{f}}_{bias,t}\}^{T}_{t=1}$.
Our contrastive loss is also implemented in the debiased module and can be formulated as follows:
\begin{equation}
    \mathcal{L}_{contras}= -\frac{1}{T}\sum_{t=1}^T \text{log}(\frac{e^{\text{score}(\bm{f}_t,\bm{f}_{debiased,t})}}{e^{\text{score}(\bm{f}_t,\bm{f}_{debiased,t})}+e^{\text{score}(\bm{f}_t,\widetilde{\bm{f}}_{bias,t})}}),
\end{equation}
where $\text{score}(\cdot)$ denotes the scoring function, of which the higher the value is, the higher the similarity between two features. Here, we utilize the cosine similarity formulated as follows:
\begin{equation}
    \text{score}(\bm{f}_1,\bm{f}_2) = \frac{\bm{f}^{\top}_1 \bm{f}_{2}}{||\bm{f}_1||\cdot||\bm{f}_{2}||}.
\end{equation}
By minimizing the contrastive loss $\mathcal{L}_{contras}$, the multi-modal features $\bm{F}$ are able to approach the debiased features $\bm{F}_{debiased}$ while avoiding the negative biased features $\widetilde{\bm{F}}_{bias}$. In this way, the features $\bm{F}$ generated by the grounding backbone have a similar representation with the debiased features $\bm{F}_{debiased}$, to some extent. Therefore, in the inference phase, we can directly remove the multi-modal debiasing branch and adopt the grounding backbone to obtain the accurate grounding results.

\subsection{Reducing the Bias by Contrastive Sampling}
\begin{figure}[t!]
\centering
\includegraphics[width=0.47\textwidth]{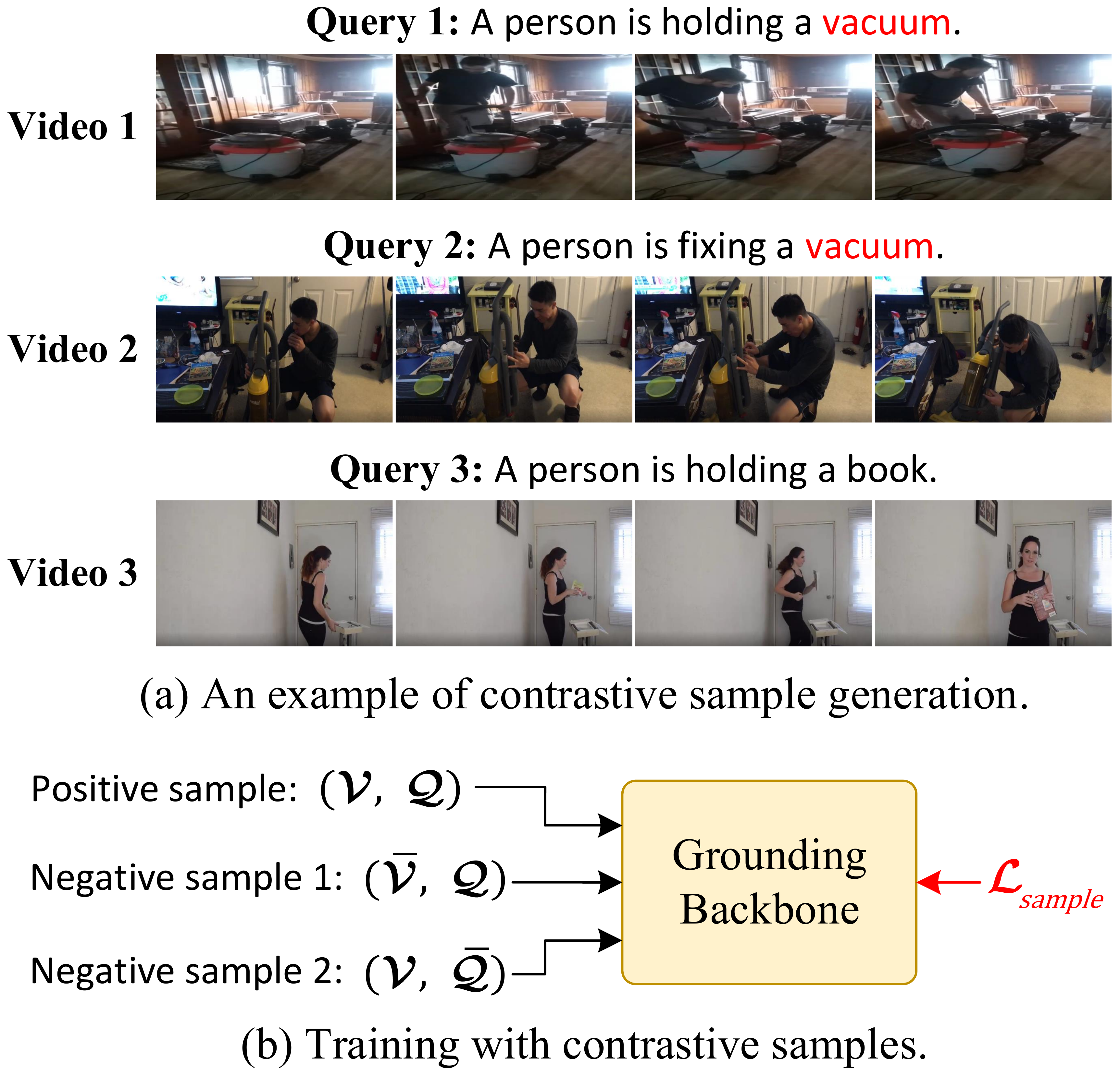}
\vspace{-10pt}
\caption{(a) The illustration of the contrasive samples which have both similar and different semantics (\textit{i.e.}, same nouns with different verbs or same verbs with different nouns). (b) The training process with contrastive samples, where $(\mathcal{V},\mathcal{Q})$ is the positive sample, $(\bar{\mathcal{V}},\mathcal{Q}),(\mathcal{V},\bar{\mathcal{Q}})$ are the negative ones.}
\vspace{-10pt}
\label{fig:contrastive}
\end{figure}

As mentioned before, existing TSG methods often learn the vision bias that only match the partial or salient semantic in the query (\textit{e.g.}, nouns or verbs) to predict the segment. 
Thus, it is necessary to increase the sensitivity of the TSG models on the complete information of both modalities. To this end, we propose a contrastive sampling strategy, in which we construct two types of negative samples for each positive sample in the dataset to assist the model training. 
Specifically, for each positive sample $(\mathcal{V},\mathcal{Q})$, we construct the negative samples by sampling the video $\bar{\mathcal{V}}$ and query $\bar{\mathcal{Q}}$ sharing the similar but different semantics to the positive one. 
In detail, given a specific query, we search its contrastive queries (different in a noun or verb) from the whole query set by making the other key words (verbs or nouns) as similar as possible.
The whole contrastive sampling process is implemented offline before training the model.
% \textcolor{red}{How to construct? The details of constructing pipeline. How can we get Query2, Query3 by Query 1.}
As shown in Figure~\ref{fig:contrastive} (a), we can collect ``Query1-Video1", ``Query1-Video2" and ``Query1-Video3" as contrastive samples. To discriminate ``Query1-Video1" as positive pair and ``Query1-Video2" as negative pair, the model needs to accurately learn the semantic ``holding" and ``fixing". Similarly, to discriminate ``Query1-Video1" as positive pair and ``Query1-Video3" as negative pair, the model needs to accurately learn the semantic ``vacuum" and ``book". We can also collect ``Query1-Video1", ``Query2-Video1" and ``Query3-Video1" as contrastive samples for the same reason.
Moreover, since the above constructed samples help to learn the specific nouns or verbs by contrastive learning, they also reduce the language bias by learning the rare query contents like ``vacuum".
During the contrastive training, as shown in Figure~\ref{fig:contrastive} (b), we randomly sample two negative samples from the above constructed sets with the supervision of the contrastive loss function $\mathcal{L}_{sample}$.

\subsection{Learning D-TSG with overall loss}
In total, our method contains three types of losses, namely, traditional TSG loss of the prediction heads (i.e., cross-entropy loss), losses for debiased feature learning, and sample perspective loss with the generated contrastive samples, which can be formulated as follows:
\begin{equation}
\begin{aligned}
    \mathcal{L} = & \mathcal{L}_{TSG} + \mathcal{L}_{bias}^1 + \mathcal{L}_{bias}^2 + \mathcal{L}_{bias}^3 + \mathcal{L}_{debias} \\
    & + \lambda_1 \mathcal{L}_{contras} + \lambda_2 \mathcal{L}_{sample},
\end{aligned}
\end{equation}
where $\mathcal{L}_{TSG},\mathcal{L}_{bias}^1, \mathcal{L}_{bias}^2,\mathcal{L}_{bias}^3, \mathcal{L}_{debias}$ are the cross-entropy loss based on the same ground-truth annotation, corresponding to the grounding backbone, three biased model in multi-modal debiasing branch, and the debiased module, respectively. $\mathcal{L}_{contras}$ is the contrastive loss in debiased module for forcing the multi-modal features in grounding backbone to near the debiased features. $\mathcal{L}_{sample}$ is the sample perspective loss with contrastive sample learning.
$\lambda_1,\lambda_2$ are hyper-parameters utilized to control the balance.

\begin{table*}[t!]
    % \small
    \centering
    \caption{Performance compared with the state-of-the-arts on the ActivityNet Caption, TACoS, and Charades-STA datasets.}
    \setlength{\tabcolsep}{1.5mm}{
    \begin{tabular}{c|cccc|cccc|cccc}
    \hline \hline
    \multirow{3}*{Method} & \multicolumn{4}{c|}{ActivityNet Captions} & \multicolumn{4}{c|}{TACoS} & \multicolumn{4}{c}{Charades-STA} \\ \cline{2-5} \cline{6-9} \cline{10-13}
    ~ & R@1, & R@1, & R@5, & R@5, & R@1, & R@1, & R@5, & R@5, & R@1, & R@1, & R@5, & R@5, \\ 
    ~ & IoU=0.5 & IoU=0.7 & IoU=0.5 & IoU=0.7 & IoU=0.3 & IoU=0.5 & IoU=0.3 & IoU=0.5 & IoU=0.5 & IoU=0.7 & IoU=0.5 & IoU=0.7 \\ \hline
    TGN \cite{chen2018temporally} & 28.47 & - & 43.33 & - & 21.77 & 18.90 & 39.06 & 31.02 & - & - & - & -\\
    CTRL \cite{gao2017tall} & 29.01 & 10.34 & 59.17 & 37.54 & 18.32 & 13.30 & 36.69 & 25.42 & 23.63 & 8.89 & 58.92 & 29.57 \\
    ACRN \cite{liu2018attentive} & 31.67 & 11.25 & 60.34 & 38.57  & 19.52 & 14.62 & 34.97 & 24.88 & 20.26 & 7.64 & 71.99 & 27.79 \\
    QSPN \cite{xu2019multilevel} & 33.26 & 13.43 & 62.39 & 40.78 & 20.15 & 15.23 & 36.72 & 25.30 & 35.60 & 15.80 & 79.40 & 45.40 \\
    CBP \cite{wang2019temporally} & 35.76 & 17.80 & 65.89 & 46.20 & 27.31 & 24.79 & 43.64 & 37.40 & 36.80 & 18.87 & 70.94 & 50.19 \\
    SCDM \cite{yuan2019semantic} & 36.75 & 19.86 & 64.99 & 41.53 & 26.11 & 21.17 & 40.16 & 32.18 & 54.44 & 33.43 & 74.43 & 58.08 \\
    GDP \cite{chenrethinking} & 39.27 & - & - & - & 24.14 & - & - & - & 39.47 & 18.49 & - & - \\
    LGI \cite{mun2020local} & 41.51 & 23.07 & - & - & - & - & - & - & 59.46 & 35.48 & - & - \\
    VSLNet \cite{zhang2020span} & 43.22 & 26.16 & - & - & 29.61 & 24.27 & - & - & 54.19 & 35.22 & - & - \\
    CMIN \cite{zhang2019cross} & 43.40 & 23.88 & 67.95 & 50.73 & 24.64 & 18.05 & 38.46 & 27.02 & - & - & - & - \\
    IVG-DCL \cite{nan2021interventional} & 43.84 & 27.10 & - & - & 38.84 & 29.07 & - & - & 50.24 & 32.88 & - & - \\
    2DTAN \cite{zhang2019learning} & 44.51 & 26.54 & 77.13 & 61.96 & 37.29 & 25.32 & 57.81 & 45.04 & 39.81 & 23.25 & 79.33 & 51.15 \\
    DRN \cite{zeng2020dense} & 45.45 & 24.36 & 77.97 & 50.30 & - & 23.17 & - & 33.36 & 53.09 & 31.75 & 89.06 & 60.05 \\
    CBLN \cite{liu2021context} & 48.12 & 27.60 & 79.32 & 63.41 & 38.98 & 27.65 & 59.96 & 46.24 & 61.13 & 38.22 & 90.33 & 61.69 \\
    MGSL \cite{liu2022memory} & 51.87 & 31.42 & 82.60 & 66.71 & 42.54 & 32.27 & 63.39 & 50.13 & 63.98 & 41.03 & 93.21 & 63.85 \\
    \hline
    \textbf{D-TSG} & \textbf{54.29} & \textbf{33.64} & \textbf{86.58} & \textbf{69.36} & \textbf{46.32} & \textbf{35.91} & \textbf{66.73} & \textbf{52.84} & \textbf{65.05} & \textbf{42.77} & \textbf{94.42} & \textbf{65.16} \\ \hline \hline
    \end{tabular}}
    \vspace{-4pt}
    \label{tab:compare}
\end{table*}

\section{Experiments}
\subsection{Datasets and Evaluation}
\noindent \textbf{ActivityNet Caption.}
ActivityNet Caption \cite{krishna2017dense} contains 20000 untrimmed videos with 100000 descriptions from YouTube. The videos are 2 minutes on average, and the annotated video clips have much larger variation, ranging from several seconds to over 3 minutes. Following public split, we use 37417, 17505, and 17031 sentence-video pairs for training, validation, and testing.

\noindent \textbf{TACoS.}
TACoS \cite{regneri2013grounding} is widely used on TSG task and contain 127 videos. The videos from TACoS are collected from cooking scenarios, thus lacking the diversity. They are around 7 minutes on average. We use the same split as \cite{gao2017tall}, which includes 10146, 4589, 4083 query-segment pairs for training, validation and testing.

\noindent \textbf{Charades-STA.}
Charades-STA is built on \cite{sigurdsson2016hollywood}, which focuses on indoor activities. 
% As Charades dataset only provides video-level paragraph description, the temporal annotations of Charades-STA are generated in a semi-automatic way.
In total, the video length on the Charades-STA dataset is 30 seconds on average, and there are 12408 and 3720 moment-query pairs in the training and testing sets, respectively.

\noindent \textbf{Evaluation.} We adopt ``R@n, IoU=m" as our evaluation metrics. The ``R@n, IoU=m" is defined as the percentage of at least one of top-n selected moments having IoU larger than m.
% During the inference, we directly construct the top-n segments by considering the summed scores of the selected start and end boundary timestamps in Equation (\ref{score1}) and (\ref{score2}).

\subsection{Implementation Details}
As for video encoding, we first utilize the $112 \times 112$ pixels shape of every frame of videos, then define continuous 16 frames as a clip and each clip overlaps 8 frames with adjacent clips. We apply C3D \cite{tran2015learning} to encode the videos on ActivityNet Caption, TACoS, and I3D \cite{carreira2017quo} on Charades-STA. 
Since some videos are overlong, we uniformly downsample the length of video feature sequences to $T=200$ for ActivityNet Caption and TACoS datasets, $T=64$ for Charades-STA dataset. 
As for sentence encoding, we set the length of word feature sequences to $M=20$, and utilize Glove embedding \cite{pennington2014glove} to embed each word to 300 dimension features. The dimension $D$ is set to 512. 
The balanced weights of $\mathcal{L}$ are set to $\lambda_1=\lambda_2=1.0$.
We train the whole model for 100 epochs with batch size of 64 and early stopping strategy.
Parameter optimization is performed by Adam optimizer with leaning rate $4\times 10^{-4}$ for ActivityNet Caption and Charades-STA and $3\times 10^{-4}$ for TACoS, and linear decay of learning rate and gradient clipping of 1.0. 
% All the experiments are implemented on a single NVIDIA TITAN XP GPU.

\subsection{Comparison with the State-of-the-Arts}
\noindent \textbf{Compared methods.}
We compare the proposed D-TSG model with
state-of-the-art TSG methods on three datasets. These
methods are grouped into two categories by the viewpoints
of proposal-based and proposal-free approach: 1) proposal-based methods: TGN \cite{chen2018temporally}, CTRL \cite{gao2017tall}, ACRN \cite{liu2018attentive}, QSPN \cite{xu2019multilevel}, CBP \cite{wang2019temporally}, SCDM \cite{yuan2019semantic}, CMIN \cite{zhang2019cross}, 2DTAN \cite{zhang2019learning}, DRN \cite{zeng2020dense}, CBLN \cite{liu2021context} and MGSL \cite{liu2022memory}. 2) proposal-free methods: GDP \cite{chenrethinking}, LGI \cite{mun2020local}, VSLNet \cite{zhang2020span}, IVG-DCL \cite{nan2021interventional}. All the results are borrowed from the reported results of their papers.

\noindent \textbf{Comparison on ActivityNet Caption.}
We compare our D-TSG model with the state-of-the-art methods on the ActivityNet Caption dataset in Table~\ref{tab:compare}. We follow the previous methods to use the same C3D features for fair comparisons. Particularly, our model outperforms the previously best proposal-based method MGSL \cite{liu2022memory} by 2.42\%, 2.22\%, 3.98\% and 2.65\% absolute improvement in terms of all metrics, respectively. It demonstrates that our method achieves better model generalization than the memory-based model MGSL by reducing the multi-modal bias. Compared to the previously best proposal-free method IVG-DCL \cite{nan2021interventional}, we also outperform them by 10.45\% and 6.54\% in terms of R@1, IoU=0.5 and R@1, IoU=0.7, demonstrating the effectiveness of reducing the bias.

\begin{figure*}[t!]
\centering
\includegraphics[width=1.0\textwidth]{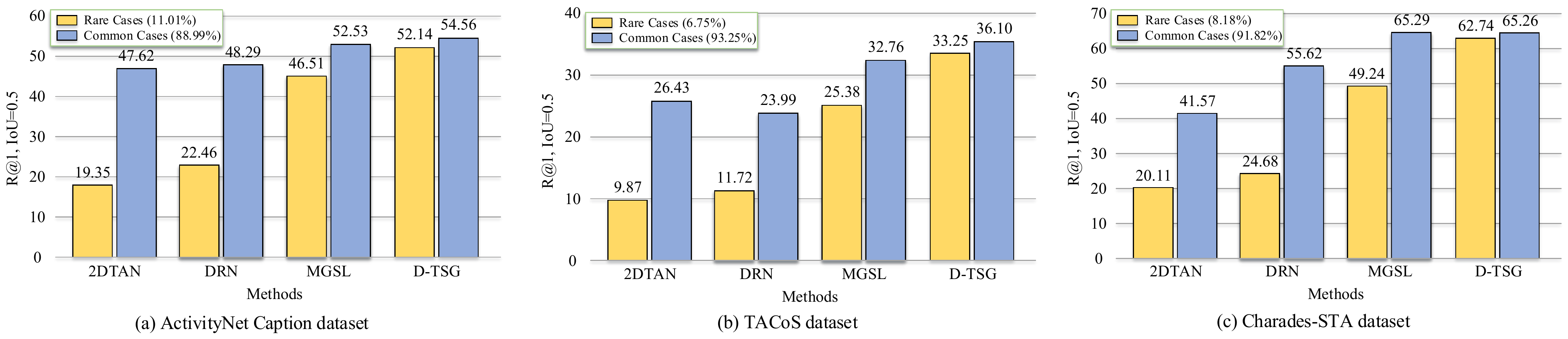}
\vspace{-16pt}
\caption{Analysis of the performance on both rare and common cases on three benchmark dataset.}
\vspace{-10pt}
\label{fig:fewshot}
\end{figure*}

\noindent \textbf{Comparison on TACoS.}
We compare our D-TSG model with the state of-the-art methods on TACoS dataset with the same C3D features in Table~\ref{tab:compare}. Particularly, our model outperforms the previously best proposal-based method MGSL \cite{liu2022memory} by 3.78\%, 3.64\%, 3.34\% and 2.71\% absolute improvement in terms of all metrics, respectively. Compared to the previously best proposal-free method IVG-DCL \cite{nan2021interventional}, we also outperform them by 7.48\% and 6.84\% in terms of R@1, IoU=0.5 and R@1, IoU=0.7, respectively.

\noindent \textbf{Comparison on Charades-STA.}
Table~\ref{tab:compare} also report the grounding comparison on Charades-STA dataset. It shows that our D-TSG model still achieves new state-of-the-art performance over all metrics.
Particularly, our model outperforms the previously best proposal-based method MGSL \cite{liu2022memory} by 1.07\%, 1.74\%, 1.21\% and 1.31\% absolute improvement in terms of all metrics, respectively. Compared to IVG-DCL \cite{nan2021interventional}, we also outperform them by 14.81\% and 9.89\% in terms of R@1, IoU=0.5 and R@1, IoU=0.7, respectively.

\subsection{Efficiency Comparison}
We evaluate the efficiency of our proposed D-TSG model, by fairly comparing its running time and model size in inference phase with existing methods on a single Nvidia TITAN XP GPU on TACoS dataset. As shown in Table \ref{tab:efficient}, it can be observed that we achieve much faster processing speeds with relatively less learnable parameters. This attributes to: 
1) The proposal-based methods (ACRN, CTRL, TGN, 2DTAN, DRN) suffer from the time-consuming process of proposal generation and proposal matching. Compared to them, our grounding backbone utilize the proposal-free grounding head, which is much more efficient.
% 2) Almost all previous methods (ACRN, CTRL, TGN, 2DTAN, DRN) directly learn the TSG model without considering the vision and language bias in the dataset, thus their performances are quite limited. Compared to them, we learn to reduce the multi-modal bias by constructing a multi-modal debiasing branch, thus enhancing the model generalization ability and leading to better grounding results.
2) The proposal-free method VSLNet (other proposal-free methods in Table~\ref{tab:compare} are not open-source or are not implemented on TACoS dataset) achieves slower speed than our D-TSG since they utilize additional query-guided highlighting module for reasoning. They have relatively smaller model size because their video and query encoders are shared weights.
However, they fail to reduce the selection bias learned from the dataset thus achieve worse performance than ours.
3) Although recent work MGSL tries to alleviate the language bias by employing a large memory bank for knowledge storing, it not only ignores the importance of vision bias and also relies on a heavy model. Different from it, our D-TSG only learns a robust grounding backbone and introduces no additional parameters during the inference. Besides, we consider reducing both vision and language bias in a single model. Therefore, our method is both effective and efficient.

\begin{table}[t!]
    % \small
    \centering
    \caption{Efficiency comparison run on TACoS dataset.}
    \setlength{\tabcolsep}{1.5mm}{
    \begin{tabular}{cccc}
    \hline \hline
    Method & Run-Time & Model Size & R@1, IoU=0.5 \\ \hline
    ACRN \cite{liu2018attentive} & 4.31s & 128M & 14.62 \\
    CTRL \cite{gao2017tall} & 2.23s & \textbf{22M} & 13.30 \\ 
    TGN \cite{chen2018temporally} & 0.92s & 166M & 18.90 \\
    2DTAN \cite{zhang2019learning} & 0.57s & 232M & 25.32\\ 
    DRN \cite{zeng2020dense} & 0.15s & 214M & 23.17 \\ 
    MGSL \cite{liu2022memory} & 0.10s & 203M & 32.27 \\
    VSLNet \cite{zhang2020span} & 0.07s & 48M & 24.27 \\
    \hline
    \textbf{D-TSG} & \textbf{0.06s} & 86M & \textbf{35.91} \\ \hline
    \end{tabular}}
    \label{tab:efficient}
    \vspace{-10pt}
\end{table}

\subsection{Analysis on the Rare Cases}
To define the few-shot contents in each TSG dataset, we follow \cite{liu2022memory} to select certain pairs of video and sentence as rare samples, which have at least one word (nouns, verbs) whose appearing frequency is less than 10. The other remained samples are treated as common samples. As shown in Figure~\ref{fig:fewshot}, we compare the performances of different TSG models on both rare and common cases. By analyzing the results, we have the conclusions as follows:
1) Previous works (2DTAN and DRN) fail to avoid the vision-language bias during training on a specific dataset, thus they tend to forget the rare cases and achieve poor performance on rare cases while converging to the distribution of common cases.
2) Although MGSL tries to alleviate the language bias by learning a memory bank with large parameters, it is still limited by the memory sizes and the convergence of the model. It also fails to reduce the vision bias.
3) Instead, our D-TSG proposes to reduce such bias by learning more robust features, which is more straightforward and effective for enhancing the model generalization ability. Therefore, we achieve much better performance on the rare cases than the other methods.

\subsection{Ablation Studies}
In this section, we perform in-depth ablation studies to evaluate the effect of each component in D-TSG on ActivityNet Caption dataset.

\begin{table}[t!]
    \small
    \centering
    \caption{Main ablation study on ActivityNet Caption dataset.}
    \setlength{\tabcolsep}{0.6mm}{
    \begin{tabular}{ccccccc|c}
    \hline \hline
    Backbone & \multicolumn{3}{c}{Biased model} & \multicolumn{2}{c}{Debiased model} & Sample & R@1, \\ 
    $\mathcal{L}_{TSG}$ & $\mathcal{L}_{bias}^1$ & $\mathcal{L}_{bias}^2$ & $\mathcal{L}_{bias}^3$ & $\mathcal{L}_{debias}$ & $\mathcal{L}_{contras}$ & $\mathcal{L}_{sample}$ & IoU=0.5 \\ \hline
    $\checkmark$ & ~ & ~ & ~ & ~ & ~ & ~ & 43.07 \\
    $\checkmark$ & ~ & ~ & ~ & ~ & ~ & $\checkmark$ & 45.64 \\
    $\checkmark$ & $\checkmark$ & ~ & ~ & $\checkmark$ & $\checkmark$ & $\checkmark$ & 47.96\\
    $\checkmark$ & ~ & $\checkmark$ & ~ & $\checkmark$ & $\checkmark$ & $\checkmark$ & 48.86\\
    $\checkmark$ & ~ & ~ & $\checkmark$ & $\checkmark$ & $\checkmark$ & $\checkmark$ & 48.62\\ \hline
    $\checkmark$ & $\checkmark$ & $\checkmark$ & ~ & $\checkmark$ & $\checkmark$ & $\checkmark$ & 51.38\\
    $\checkmark$ & ~ & $\checkmark$ & $\checkmark$ & $\checkmark$ & $\checkmark$ & $\checkmark$ & 52.11\\
    $\checkmark$ & $\checkmark$ & ~ & $\checkmark$ & $\checkmark$ & $\checkmark$ & $\checkmark$ & 50.75\\
    \hline
    $\checkmark$ & $\checkmark$ & $\checkmark$ & $\checkmark$ & $\checkmark$ & $\checkmark$ & $\checkmark$ & \textbf{54.29}  \\ 
     \hline \hline
    \end{tabular}}
    \label{tab:ablation1}
    \vspace{-10pt}
\end{table}

\noindent \textbf{Main ablation studies.}
To demonstrate the effectiveness of each component in our D-TSG, we conduct ablation studies regarding the components (\textit{i.e.}, three biased models and the debiased model in the multi-modal debiasing branch, and the contrastive sampling strategy) of D-TSG, and show the corresponding experimental results in Table \ref{tab:ablation1}. In particular, the first line represents the performance of the backbone model ($\mathcal{L}_{TSG}$), which only achieves 43.07\% in R@1, IoU=0.5. Comparing the results in other lines of this table, we have the following observations:
1) The contrastive sampling strategy constructs the negative samples to assist the model training, which promotes the model performance (refer to line 1-2 of the table),  demonstrating that our constructed negative samples are able to capture the negative biases to be removed.
2) Both biased model and debiased model co-exist to capture and debias the multi-modal bias for network learning. Specifically, each type of the biased model is able to capture the corresponding negative biases to be removed, improving the model performance (refer to line 2-5 of the table). It indicates that each type of multi-modal bias cannot be ignored during the TSG model learning.
3) Additionally, combining these three types of negative samples further promotes the model performance (refer to line 6-9 of the table).
In total, these results demonstrate the effectiveness of each component in our D-TSG method on the model performance.

\begin{table}[t!]
    \centering
    \caption{Effect of the bias identification module and the contrastive sampling strategy on ActivityNet Caption dataset.}
    \setlength{\tabcolsep}{2.0mm}{
    \begin{tabular}{c|c|cc}
    \hline \hline
    \multirow{2}*{Components} & \multirow{2}*{Changes} & R@1, & R@1,\\
    ~ & ~ & IoU=0.5 & IoU=0.7 \\ \hline
    \multirow{5}*{\tabincell{c}{Bias Identification\\ Module (BIM)}} & w/ all BIM & \textbf{54.29} & \textbf{33.64}  \\
    ~ & w/o all BIM & 50.58 & 30.37\\
    ~& w/o the BIM of $\mathcal{L}_{bias}^1$ & 53.24 & 32.81 \\
    ~& w/o the BIM of $\mathcal{L}_{bias}^2$ & 52.87 & 32.53 \\
    ~& w/o the BIM of $\mathcal{L}_{bias}^3$ & 53.15 & 32.72 \\ \hline \hline
    \multirow{4}*{\tabincell{c}{Contrastive \\ Sampling \\ Strategy}}  & w/ $(\bar{\mathcal{V}},\mathcal{Q}),(\mathcal{V},\bar{\mathcal{Q}})$ & \textbf{54.29} & \textbf{33.64} \\
    ~ & w/o $(\bar{\mathcal{V}},\mathcal{Q}),(\mathcal{V},\bar{\mathcal{Q}})$ & 51.74 & 31.03 \\
    ~ & w/o $(\bar{\mathcal{V}},\mathcal{Q})$ & 52.86 & 32.28 \\
    ~ & w/o $(\mathcal{V},\bar{\mathcal{Q}})$ & 53.15 & 32.44 \\ \hline \hline
    \end{tabular}}
    \label{tab:ablation2}
\end{table}

\begin{table}[t!]
    \centering
    \caption{Effect of different grounding backbones. Here, we directly apply our multi-modal debiasing branch and contrastive sampling strategy on the existing TSG model.}
    \setlength{\tabcolsep}{2.0mm}{
    \begin{tabular}{c|c|cc}
    \hline \hline
    \multirow{2}*{Methods} & \multirow{2}*{Changes} & R@1, & R@1,\\
    ~ & ~ & IoU=0.5 & IoU=0.7 \\ \hline
    \multirow{2}*{VSLNet} & Origin & 43.22 & 26.16 \\
    ~ & + D-TSG & \textbf{54.64} & \textbf{35.37} \\ \hline
    \multirow{2}*{CMIN} & Origin & 43.40 & 23.88 \\
    ~ & + D-TSG & \textbf{54.51} & \textbf{33.24} \\ \hline
    \multirow{2}*{DRN} & Origin & 45.45 & 24.36 \\
    ~ & + D-TSG & \textbf{55.93} & \textbf{33.12} \\ \hline
    \multirow{2}*{CBLN} & Origin & 48.12 & 27.60 \\
    ~ & + D-TSG & \textbf{57.35} & \textbf{35.98} \\ \hline \hline
    \end{tabular}}
    \label{tab:ablation3}
    \vspace{-6pt}
\end{table}

\noindent \textbf{Effect of the bias identification modules.}
The bias identification modules are devised to determine how many negative biases should be removed before the feature debiasing. To evaluate the effect of the bias identification modules in the three biased models, we conduct experiments about removing the bias identification modules in Table \ref{tab:ablation2}. We have the following observations from these results: 
1) All three bias identification modules promote the model performance individually.
2) Combining these three bias identification modules is able to further promote the model performance by a large margin.
It demonstrates the necessity and effectiveness of recognising and removing the negative biases in both language and vision modalities.

\noindent \textbf{Effect of the contrastive samples.}
As shown in Table~\ref{tab:ablation2}, we also investigate the effect of each our constructed contrastive samples. From the table, we can find that both types of the negative samples are able to improve the model performance.

\noindent \textbf{Evaluation of different grounding backbones.}
Our proposed debiasing strategy can serve as a ``plug-and-play" for existing TSG methods. As shown in Table~\ref{tab:ablation3}, we demonstrate the effectiveness of our D-TSG on different backbones on the ActivityNet Caption dataset. Specifically, we directly replace our grounding backbone with these grounding models.
From these results, we find that D-TSG is able to improve the model performance substantially regardless of the backbone, which demonstrates that our method is model-agnostic, embodying the superiority of our D-TSG. It also shows that existing TSG methods are truely suffer from the vision-language bias learned from the dataset.

\begin{figure}[t!]
\centering
\includegraphics[width=0.47\textwidth]{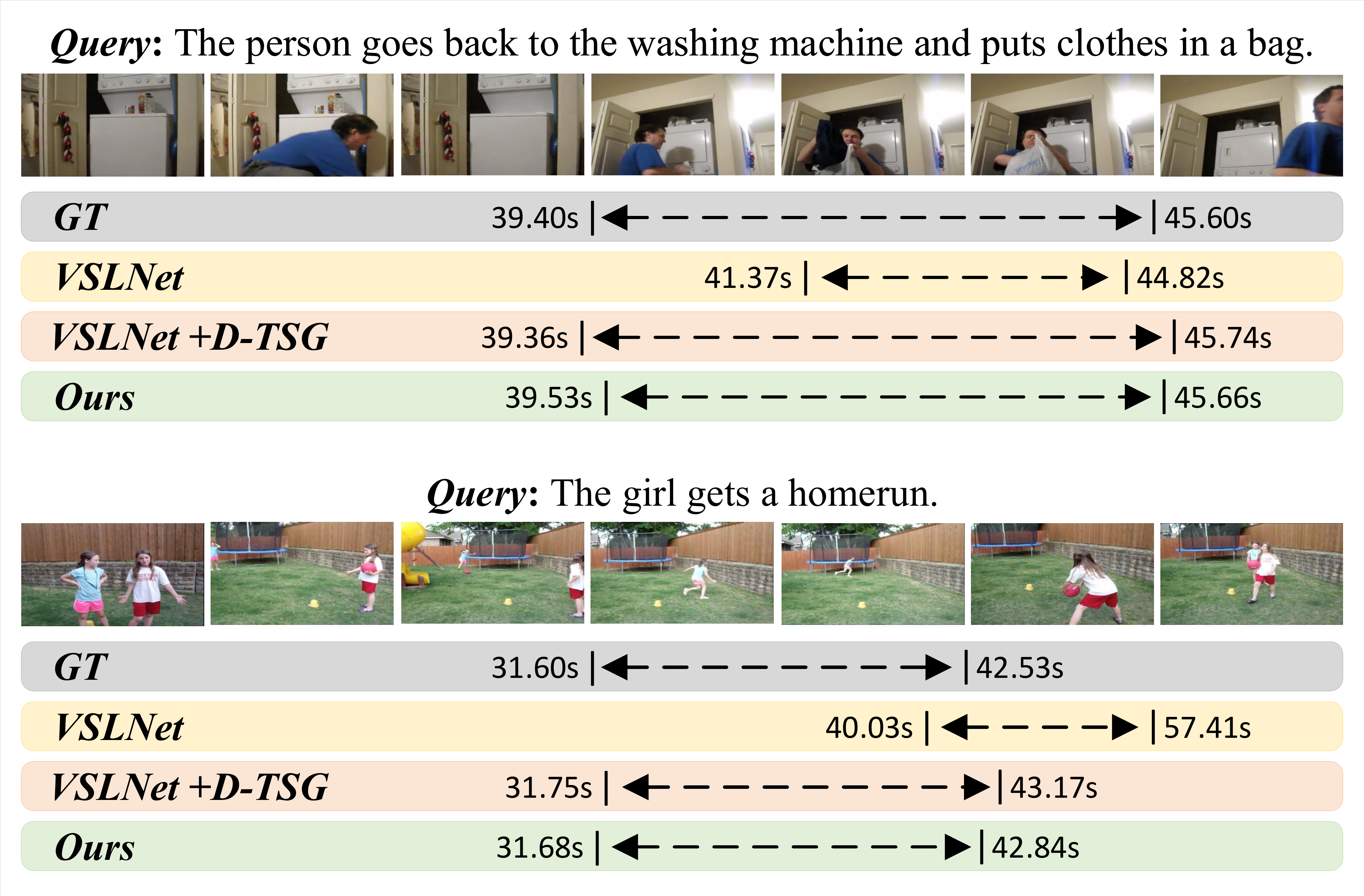}
\vspace{-2pt}
\caption{Qualitative comparison of the grounding results.}
\label{fig:result}
\vspace{-6pt}
\end{figure}

\subsection{Qualitative Results}
To qualitatively validate the effectiveness of our D-TSG model, we display two grounding examples with the vision bias and language bias in Figure~\ref{fig:result}.
The VLSNet model tends to predict the segment based on the multi-modal salient objects (vision bias) in the first example, and forget the rare words (language bias) during training in the second example, thus achieving poor grounding results. Instead, our method learns to capture and filter out both vision-language bias, thus achieving more accurate performance. Applying the D-TSG to the VLSNet also helps the VLSNet method reduce the multi-modal bias and improve the performance.

\section{Conclusion}
In this paper, we have proposed a novel method named D-TSG to overcome the negative biases in both vision and language modalities. In our D-TSG, we alleviate the negative effect of the biases from two perspectives of feature distillation and contrastive sample generation.
Specifically, from the feature perspective, we introduce a multi-modal debiasing branch to capture the different type of biases, and then devise the bias identification modules to detect and remove the true negative biases.
From the sample perspective, we construct two types of negative samples to assist the training and improve the sensitivity and generalization ability of the model.
Extensive experiments on three challenging datasets (ActivityNet Caption, TACoS, Charades-STA) show both the effectiveness and efficiency of our proposed D-TSG model.

\bibliographystyle{ACM-Reference-Format}
\bibliography{sample-base}
\end{document}